\documentclass[10pt,twocolumn,letterpaper]{article}

\usepackage{wacv}
\usepackage{times}
\usepackage{epsfig}
\usepackage{graphicx}
\usepackage{amsmath}
\usepackage{amssymb}
\usepackage[normalem]{ulem}
\usepackage{booktabs,tabularx}
\usepackage{color,subfigure,comment}

\graphicspath{ {./figs/} }

\usepackage[pagebackref=true,breaklinks=true,letterpaper=true,colorlinks,bookmarks=false]{hyperref}

\wacvfinalcopy % *** Uncomment this line for the final submission

\def\wacvPaperID{132} % *** Enter the wacv Paper ID here

\ifwacvfinal\fi
\begin{document}

%%%%%%%%% TITLE
\title{A Joint 3D-2D based Method for Free Space Detection on Roads}

\author{Suvam Patra\thanks{Equal Contribution} \\
IIT Delhi
\and
Pranjal Maheshwari$^\ast$\\
IIT Delhi
\and
Shashank Yadav\\
IIT Delhi
\and
Subhashis Banerjee \\
IIT Delhi 
\and
Chetan Arora \\
IIIT Delhi
}

\maketitle
 \thispagestyle{empty}

%%%%%%%%% ABSTRACT
\begin{abstract}
In this paper, we address the problem of road segmentation and free space detection in the context of autonomous driving. Traditional methods either use 3-dimensional (3D) cues such as point clouds obtained from LIDAR, RADAR or stereo cameras or 2-dimensional (2D) cues such as lane markings, road boundaries and object detection. Typical 3D point clouds do not have enough resolution to detect fine differences in heights such as between road and pavement. Image based 2D cues fail when encountering uneven road textures such as due to shadows, potholes, lane markings or road restoration. We propose a novel free road space detection technique combining both 2D and 3D cues. In particular, we use CNN based road segmentation from 2D images and plane/box fitting on sparse depth data obtained from SLAM as priors to formulate an energy minimization using conditional random field (CRF), for road pixels classification. While the CNN learns the road texture and is unaffected by depth boundaries, the 3D information helps in overcoming texture based classification failures. Finally, we use the obtained road segmentation with the 3D depth data from monocular SLAM to detect the free space for the navigation purposes. Our experiments on KITTI odometry dataset \cite{kitti}, Camvid dataset \cite{camvid} as well as videos captured by us validate the superiority of the proposed approach over the state of the art.
\end{abstract}

\section{Introduction}

With the rapid progress in machine learning techniques, researchers are now looking towards autonomous navigation of a vehicle in all kinds of road and environmental conditions. Though many of the problems in autonomous driving look easy in the sanitised city or highway environments of developed countries, the same problems become extremely hard in cluttered and chaotic scenarios particularly in developing countries. Detection of free road space or drivable area on road is one such problem, where many state-of-the-art techniques work successfully when the roads are well maintained and boundaries are clearly marked but fail in the presence of potholes, uneven texture and roads without well marked shoulders \cite{guo_2012,camvid}.

\begin{figure}
\begin{center}
\subfigure[]{\includegraphics[width=0.44\linewidth,height=0.1\textheight]{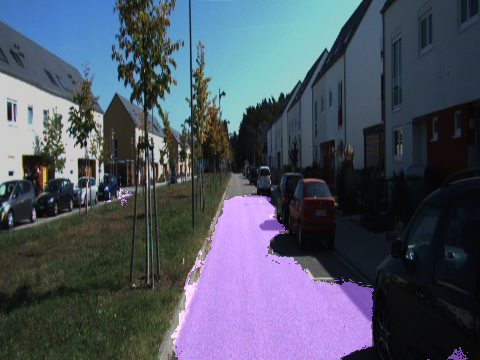}}
\subfigure[]{\includegraphics[width=0.44\linewidth,height=0.1\textheight]{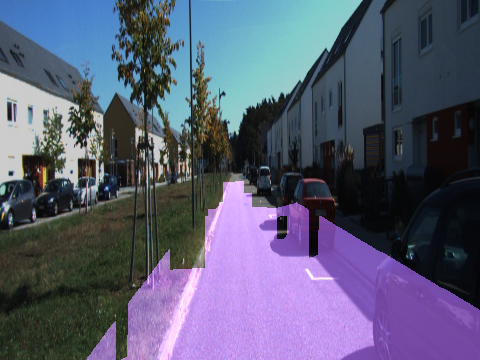}}
\\
\subfigure[]{\includegraphics[width=0.44\linewidth,height=0.1\textheight]{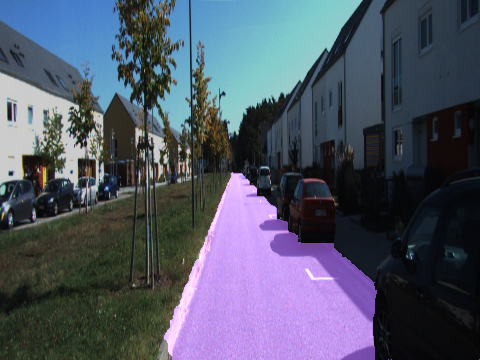}}
\subfigure[]{\includegraphics[width=0.44\linewidth,height=0.1\textheight]{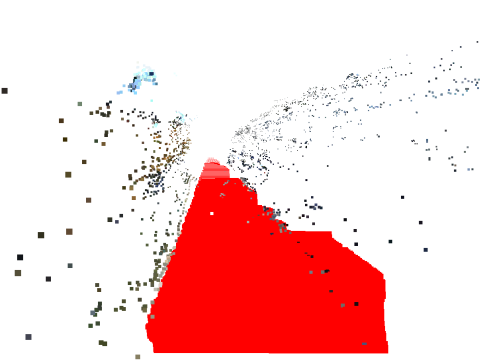}}
\end{center}
\caption{Free road space detection is an important problem for the driving assistance systems. However, (a) 2D image based solutions such as SegNet \cite{segnet15} often fail in the presence of non uniform road texture. (b) On the other hand, methods using 3D point cloud fail to identify fine depth boundaries with pavement. (c,d) The proposed technique uses both 2D and 3D information to obtain state of the art detection results both in 2D image space (c) and in 3D world coordinates (d).}
\label{fig:motivation}
\end{figure}

The free road space detection (hereinafter referred to as `free space' detection without the word `road') has been a well studied problem in the field of machine vision. Based on the modality of input data, the methods for free space detection can be broadly classified into  image based methods, 3D sensor based methods, and hybrid methods which utilise both (3D and 2D data). Purely image based 2D methods use low-level cues such as color or a combination of color and texture \cite{alvarez_2011, sotelo_2004, kim, tan, Lombardi_2005} and model the problem as a road segmentation/detection problem. With the tremendous success of deep networks in other computer vision problems, researchers have also proposed learning based methods for this problem \cite{segnet15,long_2015}.

Nowadays, depth sensors like LIDAR, RADAR, and ToF (time of flight) are readily available giving real time depth data. Methods like \cite{Fernandes_lidar} use points clouds obtained from such sensors to detect the road plane and the 3D objects present. Use of extra sensors makes the problem easier by providing real time depth cues, but they also require complex data fusion and processing from multiple sensors, thereby increasing the hardware and computational costs. Also, this kind of computation is difficult on board in autonomous cars because, in addition to detecting the freespace, the computations for controlling the motion of the car are by themselves quite complex.

The hybrid methods for free space detection \cite{Hu_multimodal,camvid} often use depth data by projecting it to 2D image space and then using it in conjunction with color and texture to obtain the road and free space detection in 2D.

\begin{figure*}[t]
\centering
\includegraphics[width=0.90\linewidth]{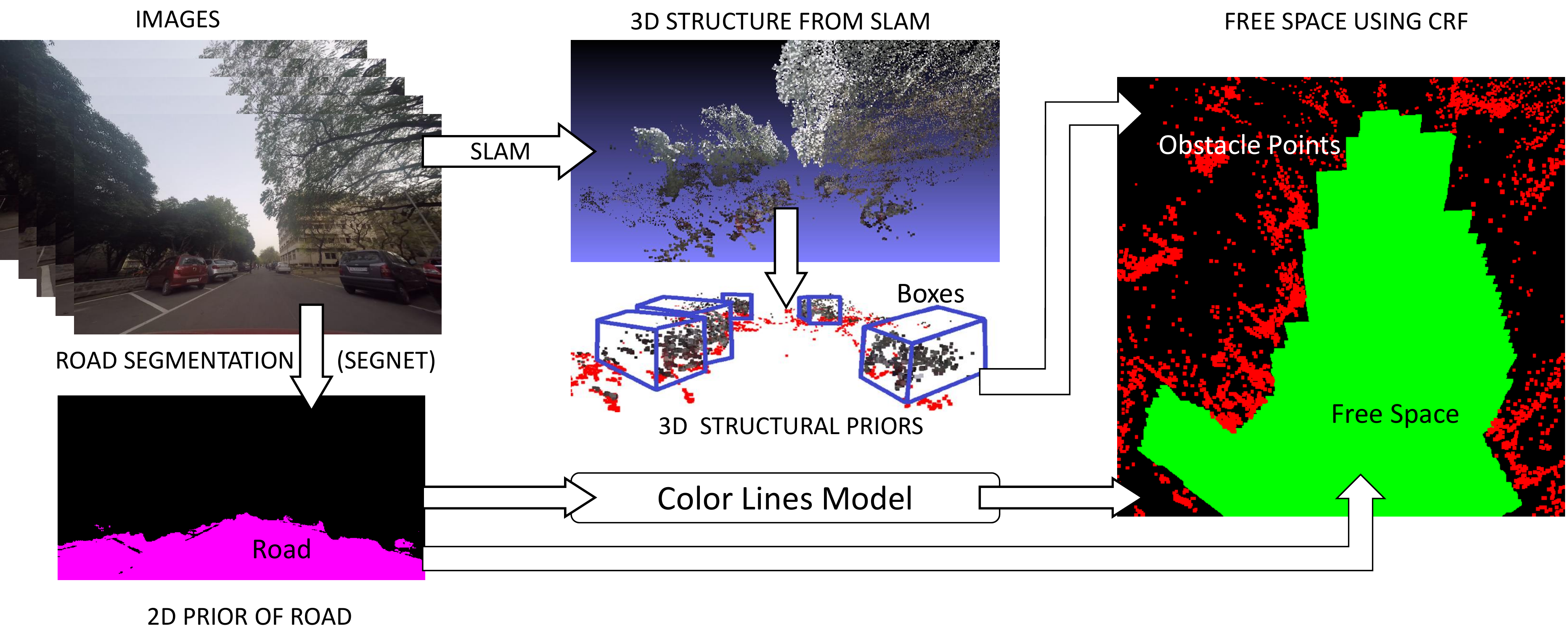}
\caption{Overall framework of our method}
\label{fig:framework}
\end{figure*}

\paragraph{Our Contributions:} We leverage complementary strengths of 2D and 3D approaches for the free space detection problem. Most of the 2D segmentation techniques fail on the non-uniform texture of the road such as those arising from potholes, road markings, and road restoration. 3D techniques though invariant to uneven textures suffer from lack of resolution on fine depth boundaries. This has motivated the proposed joint 2D-3D based approach, where:
\begin{itemize}
\item We generate higher order depth priors such as road planes and bounding boxes, from sparse 3D depth maps generated from monocular SLAM.
\item Instead of projecting sparse 3D points to 2D, we project these dense higher order priors to the 2D image, which leads to transfer of 3D information to large parts of the image.
\item The dense prior from 3D are used in conjunction with per pixel road confidence obtained from SegNet \cite{segnet15} and color lines prior \cite{color_lines} in a CRF formulation to obtain robust free space segmentation in 2D.
\item The road pixel segments in 2D are back projected and their intersection with the estimated 3D plane is used for detecting free space in 3D world coordinates.
\end{itemize}
We show qualitative as well as quantitative results on benchmarks KITTI odometry dataset \cite{kitti}, Camvid dataset \cite{camvid} and also on videos captured by us in unmarked road conditions. We give an example result of our technique in Figure \ref{fig:motivation} and the overall framework in Figure \ref{fig:framework}.

\section{Related work}

The free space detection techniques are classified on the basis of their modality of input into: 2D (image based), 3D (3D structure based) or hybrid (both). We review each of these classes below.

\paragraph{2D Techniques}

Purely image based methods that use low-level cues such as color \cite{alvarez_2011, sotelo_2004, kim, tan} or a combination of color and texture \cite{Lombardi_2005} have been used to model free space detection more like a road segmentation/detection problem. Siagian \etal \cite{mobile_robot} use road segment detection through edge detection, and voting of vanishing points. Chen \etal \cite{Chen_monocular_object_detect} approximate the ground plane and then put object proposals on the ground plane by projecting them on the monocular image. They formulate an energy minimization framework using several intuitive potentials encoding semantic segmentation, contextual information, size and location priors and typical object shapes. No 3D information is used. These methods work well for well-marked and sanitised roads in absence of depth cues. Methods like \cite{segnet15,long_2015} model  road segmentation as a texture learning problem. It may be noted that 2D methods cannot directly detect free space for car movement and can only be used as priors for drivable free space modelling.

\paragraph{3D Techniques}

Hornung \etal \cite{octomap} use hierarchical Oct-trees of voxels where a voxel is labelled free or occupied based on its detected occupancy. Kahler et al. \cite{kahler_HierarchicalVoxel} model it similarly using hashing of 3D voxels. They use hashing to store occupancy of each of these voxels at variable resolution. These class of methods is more suitable for navigation of drones where occupancy in 3D is required to control the 3-dimensional motion of the drones. These 3D techniques depends on the sparse 3D points produced by LIDAR, SONAR, ToF sensors which lack the resolution to differentiate finer details e.g. roads from pavements, potholes, etc. In this paper we propose to use additional cues from 2d images to overcome the limitation.

\paragraph{Hybrid Techniques}

Hybrid methods use depth data indirectly by computing depth using stereo cameras/LIDAR. Methods like \cite{VitorLVF13, Lategahn_occupancy, OnigaN10_densestereo, broggi_obstacle2013, BerniniBCPS14} model roads using stereo camera setups. Using a calibrated stereo rig, detailed depth maps can easily be estimated as a first step. In the second stage, these methods use the dense stereo depth maps to calculate free space in 3D. Methods like \cite{Hu_multimodal} use LIDAR data feed as input and then use images to distinguish road boundaries.

Free space detection can indirectly be modelled as an obstacle detection problem as well, wherein free space is determined as a by-product of detecting obstacles in 3D. The method described in \cite{3dopNIPS15} model this problem using 3D object proposals by fitting cuboids on objects with the help of stereo cameras. A comprehensive survey of obstacle detection techniques based on the stereo vision for intelligent ground vehicles can be found in \cite{BerniniBCPS14}.

The works closest to ours are by \cite{camvid} and \cite{Hu_multimodal}. Broswtow \etal \cite{camvid} assume only a point cloud as input which is first triangulated and then projected onto the image plane. The features computed from surface normals are used to segment the 2D image. Similarly, Hu \etal \cite{Hu_multimodal} use point cloud from LIDAR, identify points on the ground plane, project these points to the image plane and then do image segmentation based upon the projected points. Both the methods are similar in the sense that 3D points on the ground plane are projected to the image plane followed by the image segmentation. Since the 3D points are sparse, the cues transferred to the image plane are sparse too. Also, no freespace information can be derived from 2D segmentation output. We, on the other hand, compute and project 3D priors such as planes and cuboids to the image, leading to the dense 3D context transfer. Unlike the two approaches, we output drivable space both in the 2D image plane as well as in 3D world coordinates.

\section{Background}

In this section, we briefly describe the concepts necessary for understanding of our methodology.

\paragraph{Structure Estimation}

The pose of  a camera  w.r.t a global frame of reference is denoted by a $\mathrm{3 \times 3}$ rotation matrix $\mathrm{\textbf{R} \in SO(3)}$ and a $\mathrm{3 \times 1}$ translation direction vector $\textbf{t}$. Structure-from-Motion (SfM) simultaneously solves for these camera poses and the 3D structure using pairwise estimates. The pairwise poses can be estimated from the decomposition of the $\textit{essential matix}$ $E$ which binds two views using pairwise epipolar geometry such that: $E= [\textbf{t}]_\times\textbf{R}$ \cite{mvg,nisterfive}. Here $[\textbf{t}]_\times$ is a skew-symmetric matrix corresponding to the vector $\textbf{t}$. These pairwise poses then can be used to estimate the structure and camera parameters simultaneously using SfM based Simultaneous Localization And Mapping (SLAM) framework \cite{lsd-slam,ptam,orb-slam,dtam,RSLAM2017}. We use the SLAM method described in \cite{RSLAM2017} where the camera poses in small batches are first stabilized using motion averaging, followed by a SfM refinement using global bundle adjustment for accurate structure estimation.

\paragraph{Illumination Invariant Color Modelling}
%\label{sec:color-lines}

The road detection techniques based on colors often fail to adjust to the varying illumination conditions such as during different times of the day. Color lines \cite{color_lines} model predicts how the observed color of a surface changes with change in illumination, shadows, sensor saturation, etc. The model discretizes the entire histogram of color values of an image into a set of lines representing various regions which have similar color values. The entire RGB space is quantized into bins using discrete concentric spheres centered at origin. The radius increases in multiples of some constant integer. The volume between two consecutive spheres is defined as a bin. Bin $k$ contains the set of RGB values of image points lying in the volume between $k^{th}$ and the ${k+1}^{th}$ concentric spheres. This representation also gives a metric for calculating distance between every pixel and each color line as shown in \cite{color_lines}. Using such a distance metric helps in improved texture modelling under varying illumination conditions.

\paragraph{Image Segmentation}

Image based segmentation techniques are semantic pixel-wise classifiers which label every pixel of an image into one of the predefined classes. The approach has been widely used for the road segmentation as well. SegNet \cite{segnet15} is one such method which can segment an image into 12 different classes which include road, cars, pedestrians, etc. SegNet uses a deep convolutional neural network with basic architecture identical to VGG16 network \cite{vgg16}. The training for the SegNet has been done on well marked and maintained roads and the pre-trained model fails to adapt to the difference in texture as observed in unmarked roads from other parts of the world. Road markings and other textures present on the road such as those arising from patchwork for the road restoration also interfere with the SegNet output. In the proposed approach we make up for these weaknesses by using depth cues along with color lines based cost in our model. It may be noted that we use SegNet because of its ability to learn road texture efficiently. Other similar techniques (based on deep learning or without) could have been equivalently used without changing the proposed formulation.

\section{Proposed Approach}

The output of a SLAM (Simultaneous localization and mapping) approach is the camera pose and the sparse depth information of the scene in the form of a point cloud. We use the point cloud generated from \cite{RSLAM2017}, referred to as Robust SLAM in this paper, as input to our algorithm. Similar to the choice of SegNet, we use Robust SLAM because of its demonstrated accuracy in the road environment. The point cloud generated from any other SLAM or SfM algorithm can also be used in principle. We use this 3D point cloud to generate higher order priors in the form of a ground plane (for the road) and bounding boxes or rectangular parallelepipeds (for obstacles like cars, pedestrians, etc). The overview of our method is shown in Figure \ref{fig:framework}.

\subsection{Generating Priors from the 3D Structure}\label{sec:3d_priors}

\begin{figure}[t]
\centering
\includegraphics[width=0.85\columnwidth]{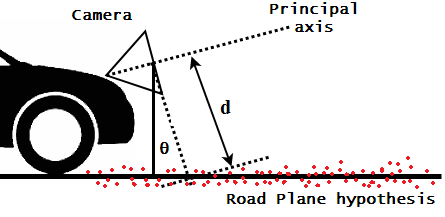}
\caption{Road plane detection from point cloud. Road points are marked in red. The road plane is fitted on the basis of the angle $\theta$ between road plane and principal axis and distance $d$ of the principal axis from the ground plane}
\label{fig:road_prior}
\end{figure}

The road plane detection is facilitated by using 3D priors estimated from the input 3D point cloud. We search for the road plane in a parametric space using a technique similar to Hough transform \cite{ballard_hough}. Our parametric space consists of the distance of the plane from camera center and the angle of the plane normal with the principal axis of the camera. The equation of the plane in the parametric space of $d$ (the distance of the road plane from the principal axis) and $\theta$ (angle of the principal axis with the road plane) is given as:
\begin{equation}
z ~ sin(\theta) - y ~ cos(\theta) = d ~ cos(\theta)
\end{equation}
For more details refer to Figure \ref{fig:road_prior}. The estimation process assumes that the height of the camera and the orientation with respect to the road remains fixed. We use wheel encoders, for scale correction of the camera translations and the point cloud obtained from SLAM to a metric space. The scale of the distance obtained from the encoders is used to correct the camera translation scales, which in turn automatically scales the point cloud as well to the metric space. This is necessary as it helps in the parametric plane fitting with a known initialization of $d$ based on the known height of the camera.

Once the road plane is estimated, we cluster the points above the road plane. We have used $K$-means clustering \cite{kmeans} with large enough value of $K$. As it will become evident, over clustering that may result because of this choice is less problematic than under clustering because our objective is free-space detection and not obstacle detection, where finding object boundaries are more important. For example, even if we fit two clusters over a single car, it does not affect the free space estimation as long as the union of the two clusters covers the whole area occupied by the car.

For each cluster, we use principal component analysis (PCA) \cite{pca} to find two largest eigenvectors perpendicular to the normal of the road plane. We fit enclosing boxes on each of the clusters along these orthogonal eigenvectors and the road normal. This representation is directly used in our model as the 3D priors (shown in Figure \ref{fig:framework}).

\begin{figure*}[t]
\begin{center}
\includegraphics[width=0.95\linewidth]{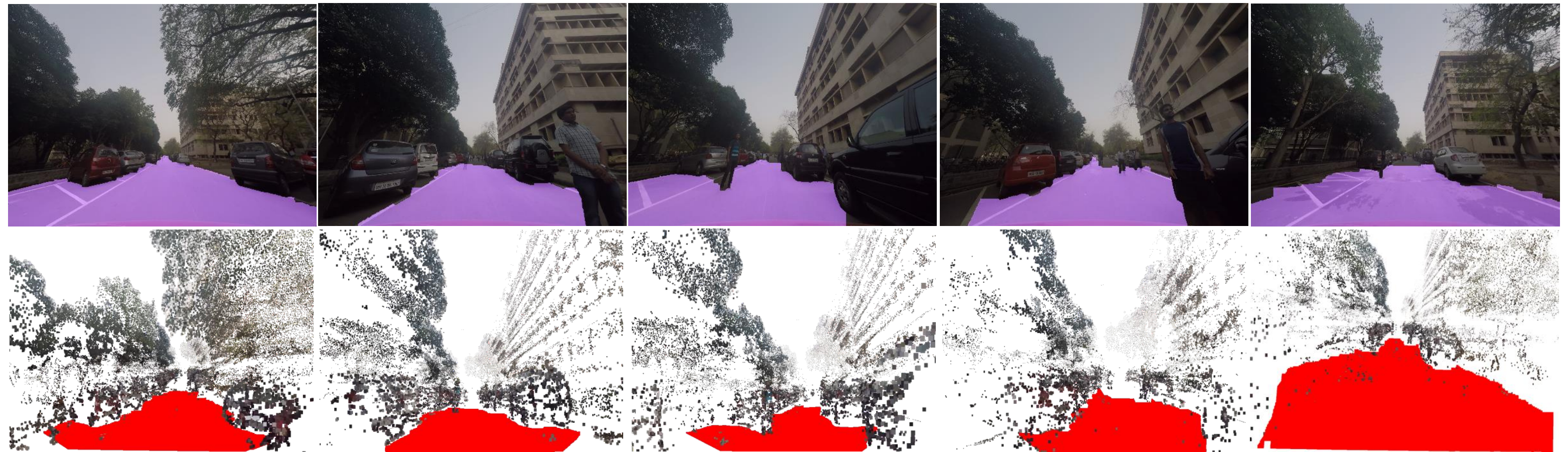}
\end{center}
\caption{Our result on few images from a sequence taken by us in the University campus. The first row shows the 2D projection of the free space on the image, second row shows depiction of detected free space in 3D on the sparse point cloud obtained using SLAM. }
\label{fig:ours_3d}
\end{figure*}
\subsection{2D Road Detection: Problem Formulation} \label{sec:2droad}

In the first stage, we detect the road pixels in 2D image space. We formulate the problem as a 2-label energy minimization problem over image pixels. Here the two labels are $R$ and $\backslash R$ for `road' and `not road' respectively. The confidence from the 3D estimation is transferred by projecting the estimated road plane onto the image using camera parameters. Note that even though the point cloud obtained from SLAM is sparse, the 3D plane projection leads to transfer of 3D information from large parts of the road to the image. Further, it is possible that 3D plane erroneously suggests the adjoining pavement or other low lying areas as the road. This is not problematic since we combine this confidence with the 2D cues ultimately in a joint CRF (Conditional Random Field) \cite{c111,Sutton_crf} formulation which we describe below.

\paragraph{Cost from 3D Priors:}

For each pixel $p$ in the image we shoot a ray $R_p$ and compute its first intersection (in forward ray direction only) with available 3D objects (road plane or fitted 3D boxes). This defines an indicator function for the pixel for the current frame as follows:
\begin{equation}
I(p) =
  \begin{cases}
    1 & \text{if } R_{p}  \text{ intersects road plane first}\\
    0 & \text{otherwise}\\
  \end{cases}
\end{equation}

We use the indicators to define the 3D component of the data term corresponding to pixel $p$ as follows. We associate a cost $\omega_1$ with labeling pixel $p$ as road ($p \in R$) in the CRF, if the ray does not intersect the road plane. The cost is zero  if the pixel gets labeled as `not road' ($p \in \backslash R$). Similarly, if the ray intersected the road plane, we assign a cost $\omega_2$ if the pixel is labeled `not road' ($p \in \backslash R$) and the cost is zero if labeled `road' ($p \in R$ ). Mathematically:
\begin{equation}
D_{1}(p) =
\begin{cases}
\omega_1 (1-I(p)) & \text{if $p \in R$} \\
\omega_2 I(p)  & \text{if $p \in \backslash R$}.
\end{cases}
\label{data_indicator}
\end{equation}

For temporal consistency of 3D priors, we also use the plane and objects fitted in the point cloud corresponding to the previous frame by transferring them to the current frame using the pairwise camera pose obtained using SLAM. We define another data cost $D_{2}(p)$ similar to Equation \ref{data_indicator} where the per-pixel ray intersection of the current image is computed w.r.t the transferred 3D priors (corresponding to the previous frame).

\paragraph{Cost from SegNet:}

The pixel $p$ is also classified by SegNet as road or non-road. The corresponding data term utilizing the SegNet classification probabilities is formulated as:
\begin{equation}
D_3(p) =
  \begin{cases}
    \omega_3 \max S_{\backslash R}(p) & \text{if $p\in R$} \\
    \omega_3 S_{R}(p) & \text{if $p \in \backslash R$}.
  \end{cases}
  \label{data_segnet}
\end{equation}
Here $S_{R}(p)$ is the probability (softmax) outputted by SegNet for label $R$ (road)  and $\max S_{\backslash R}(p)$ denotes the maximum probability among all the labels which are not road ($\backslash R$). In both the cases we have associated a weight of $\omega_3$ with the respective costs (road or not road).

\begin{figure*}
\begin{center}
\includegraphics[width=0.95\linewidth]{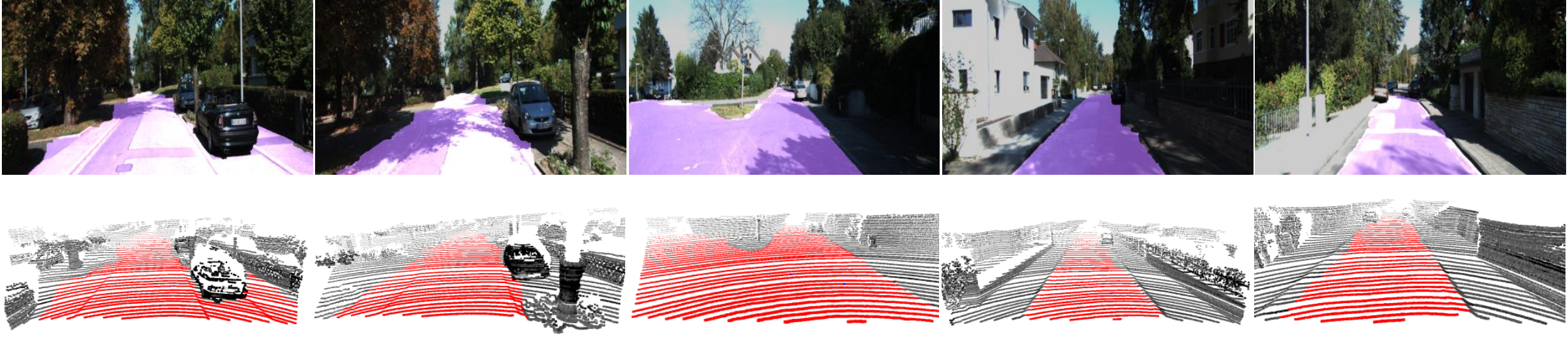}
\end{center}
\caption{Our result on some sample images from sequence 02 and 03 from the KITTI odometry data set \cite{kitti}, the free space for each of the images is shown in 2D in the first row(overlaid in pink) and in 3D in the second row as a depiction of detected free space overlapped on the corresponding ground truth depth scan obtained from LIDAR marked in red.}
\label{fig:kitti_3d}
\end{figure*}

\paragraph{Smoothness Term for the CRF:}

We use scores from the color lines model to compute smoothness of label assignments.  The standard SegNet model completely fails to learn the road texture model under varying illumination such as during different times of the day. The varying surface conditions such as from  potholes or road restoration also create problems. We have used color lines model to obtain illumination invariant color representations and used it to generate a confidence measure for each pixel being labeled as $R$ (road).

We use SegNet to bootstrap the color lines model. We use the pixels labeled as $R$ (road) by SegNet to create bins in the RGB color space for creating the color lines model as detailed out in Section \ref{sec:color-lines}. For each bin, a representative point is computed as the mean of the RGB values of the road pixels in that bin. Along with it, we also calculate the variance of the RGB colors in that bin. Henceforth, this mean and variance are used to characterize the bin.

For each pixel $p$ with RGB value $x$, we first compute the bin it lies in. We then compute the probability of that pixel being on the road using a Gaussian distribution with the characteristic mean and variance of that bin computed as described above. In case there is no representative point in a bin, as SegNet may have missed some road points because of illumination differences, we use the color lines model to extrapolate the line to that bin. This probability score from the color lines model and the line interpolation predicts with high probability such pixels to be from the road model.

We use a 4 pixel neighbourhood ($\mathcal{N}$) for defining the smoothness term. We compute the capacity of an inter-pixel edge between pixels $p$ and $q$ as:
\[
V(p,q) = P_R(p|i) * P_R(q|j) + P_{\backslash R}(p|i) * P_{\backslash R}(q|j).
\]
Here we assume that pixels $p$ and $q$ belongs to bin $i$ and $j$ respectively. $P_R(p|i)$ and $P_{\backslash R}(p|i)$ represent the scores of pixel $p$ for originating from road ($R$) and background/non-road ($\backslash R$) respectively. The formula ensures that the edge weight between the two neighboring pixel is strong if both of them have high probability of being road or high probability of being not road.

The complete energy function for the CRF is thus defined as:
\begin{equation}
E = \sum_{\forall p} \left( D_1(p) + D_2(p) + D_3(p) \right) + \sum_{\forall p, q \in \mathcal{N}} V(p,q)
\end{equation}
We find the labeling configuration with maximum a posteriori probability using Graph Cuts \cite{c111}.

\subsection{Free Space in World Coordinates}

Most of the contemporary approaches give free space information in 2D coordinates which have limited utility for navigation. However, in the proposed approach, we output free space information in 3D world coordinates as well. For each pixel detected as the road by our method, we shoot back a 3D ray and compute its intersection with the 3D road plane as described in section \ref{sec:3d_priors}. The set of such 3D intersections gives us the desired free space information in the 3D world coordinates.

\begin{figure*}
\begin{center}
\includegraphics[width=0.95\linewidth]{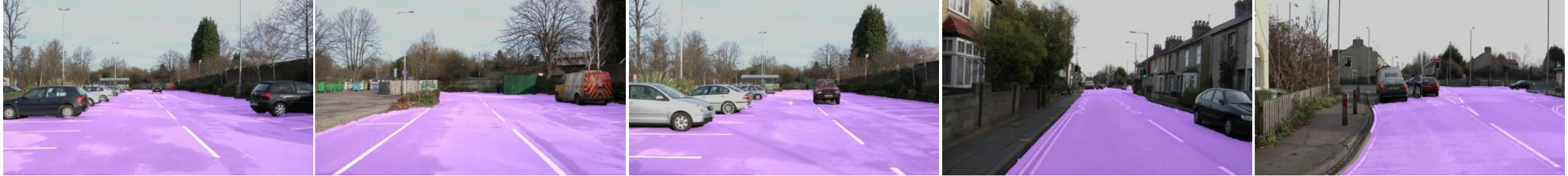}
\end{center}
\caption{Our result on some sample images from Camvid dataset \cite{camvid}, the free space for each of the images is shown in 2D (overlaid in pink).}
\label{fig:camvid_3d}
\end{figure*}

\section{Experiments and Results}

In this section, we show the efficacy of our technique on the publicly available KITTI \cite{kitti}, and Camvid \cite{camvid} datasets as well as on some videos captured by us. Our videos were obtained from a front facing GoPro \cite{gopro} camera mounted on a car recording at 30 fps in a wide angle setting.

We have implemented our algorithm in C++. All the experiments have been carried out on a regular desktop with Core i7 2.3 GHz processor (containing 4 cores) and 32 GB RAM, running Ubuntu 14.04.

Our algorithm requires the intrinsic parameters of the cameras for SfM estimation using SLAM \cite{RSLAM2017}. For the sequences obtained from public sources, we have used the calibration information provided on their websites. For our own videos, we have calibrated the GoPro camera.

It may be noted that our algorithm can work with both pre-computed point clouds and with those computed online. For pre-computed clouds, we need to relocalize the camera and then use the proposed technique. This can make the proposed approach faster at the expense of adaptability to changes in the scene structure that might occur. In our experiments, we have used the latter and use Robust SLAM \cite{RSLAM2017} to generate point cloud on the fly. In our experiments, we have chosen $\omega_1$, $\omega_2$ and $\omega_3$ to be 0.9, 0.9 and 1.0 respectively (Section \ref{sec:2droad}) .

We are able to process each key frame in 0.5 sec considering one key frame taken every 15-20 frames for a 30fps video. Note that the computation can be made faster using parallel computations employing a GPU (not done in our experiments). For 2D free space computation step, the major bottleneck is SegNet computations, which takes about 300 ms for one key frame computation. We understand that there are many newer models like \cite{maskRCNN} which can potentially speed up the computation.

\subsection{Qualitative Results}

In Figure \ref{fig:ours_3d} we show the results of detected free space using our method on a video obtained by us. We show the detected free space both in 2D (projection on the image in pink) and in 3D (as a plane shown in red) on the sparse point cloud obtained from SLAM.

We also test our algorithm qualitatively on some of the sequences from the challenging KITTI odometry dataset \cite{kitti}. Unlike our case with only monocular video information, the dataset also provides the ground truth of point cloud in the form of LIDAR data for each frame. We use the LIDAR point cloud for visualization purposes. Note that, we do not use the LIDAR point cloud for the free space detection in our algorithm where the point cloud from Robust SLAM is used.

In Figure \ref{fig:kitti_3d}, we show the free space estimated through our algorithm in 2D (projected on the image in pink) and also in 3D overlaid on the LIDAR data (red) for some sample images from the KITTI dataset. Figure \ref{fig:camvid_3d} demonstrate similar results on the Camvid dataset. Here we show only 2D images because of unavailability of LIDAR data for the overlaying and visual comparison.

\begin{figure*}
\begin{center}
\includegraphics[width=0.95\linewidth]{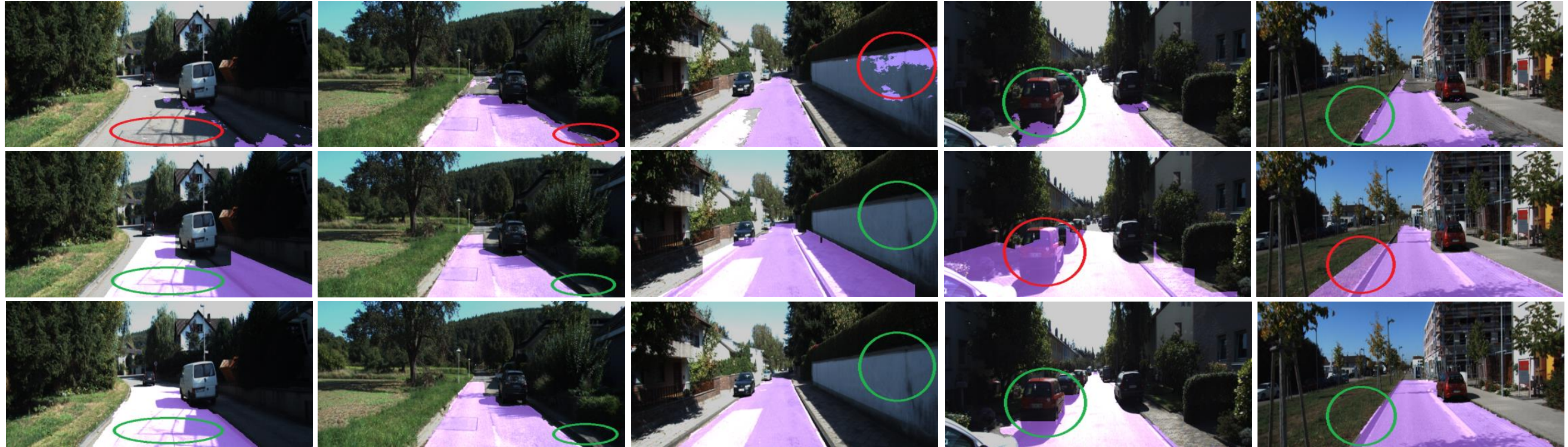}
\end{center}
\caption{Some examples from KITTI odometry dataset \cite{kitti}) where our method is able to fix errors in free road space detection from only 2D image based priors (first row) or 3D depth based priors (second row). Our results are in the third row (corrected areas marked in green).}
\label{fig:kitti_seg}
\end{figure*}

\begin{figure*}
\begin{center}
\includegraphics[width=0.95\linewidth]{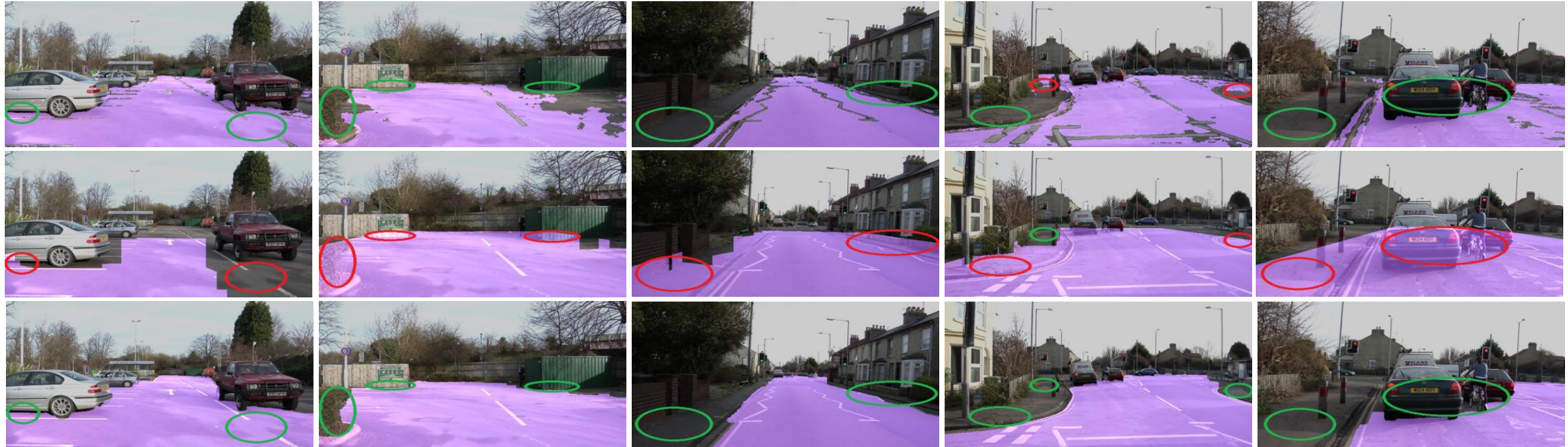}
\end{center}
\caption{Similar analysis as shown in Figure \ref{fig:kitti_seg} but on samples from the Camvid dataset \cite{camvid}}
\label{fig:kitti_depth}
\end{figure*}

One of our important claims is that the joint 2D-3D formulation is able to mitigate the shortcomings of 2D or 3D alone. We demonstrate this qualitatively on some images taken from the KITTI odometry dataset \cite{kitti} in Figure \ref{fig:kitti_seg}. In the first row of Figure \ref{fig:kitti_seg} we show some typical cases where using 2D information only for free space detection sometimes leads to anomalies. The areas of discrepancies are marked in red. Using 3D only is also not sufficient often and leads to failures on low depth differences as shown in row 2 with discrepancies marked in red. Using both 2D and 3D information as suggested in the proposed approach helps fix many of these errors as shown in row 3 of Figure \ref{fig:kitti_seg} with corrected areas marked in green. Figure \ref{fig:kitti_depth} shows similar results for some images from the Camvid dataset \cite{camvid}.

Since the code for \cite{Hu_multimodal} is not publicly available we are unable to compare. However, we note that their scheme is conceptually similar to switching off the 2D prior in our proposed CRF formulation, and is likely to perform inferior. Badrinarayanan \etal \cite{segnet15} have shown that their technique, SegNet, gives better performance than \cite{camvid} (see Table I in \cite{segnet15}). Therefore, we compare only with SegNet as the current state of the art.

\subsection{Quantitative Evaluation}

We perform the quantitative evaluation on the Camvid and KITTI odometry benchmark dataset. For this purpose, we used three different standard pixel-wise measures used in literature i.e. F val, precision and recall as defined in Table \ref{tableEval}.
\begin{table}[htp]
\centering
\begin{tabular}{ll}
\toprule[1.5pt]
\bf Pixel-wise Metric & \bf Definition  \\ \hline
Precision &  $\frac{TP}{TP+FP}$ \\
Recall &  $\frac{TP}{TP+FN}$  \\
Fval (F1-score) & 2$\frac{Precision*Recall}{Precision+Recall}$ \\
 \bottomrule[1.5pt]
 \end{tabular}
\vspace{0.2cm}
  \caption{Performance evaluation metrics} \label{tableEval}
\end{table}
\noindent Here $TP$ is the number of correctly labeled road pixels, $FP$ is the number of non-road pixels erroneously labeled as road and $FN$ is the number of road pixels erroneously marked as non-road. Precision and Recall provide different insights into the performance of the method: Low precision means that many background pixels are classified as road, whereas low recall indicates failure to detect the road surface. Finally, F value i.e. F1-measure (or effectiveness) is the trade-off using weighted harmonic mean between precision and recall.

We compare the performance of proposed approach against SegNet \cite{segnet15} which uses only 2D information and our formulation without using 2D priors (as an indicator of performance using depth only). On the  Camvid benchmark dataset \cite{camvid} our system outperforms both SegNet \cite{segnet15} (Image based) and depth based method in all three scores as shown in Table \ref{tableCamvidtest}. Here the precision is almost the same because SegNet has been trained on the Camvid dataset.

\begin{table}[htp]
\centering
\begin{tabular}{lccc}
\toprule[1.5pt]
\bf Method & \bf F val & \bf Precision & \bf Recall  \\ \hline
2D (SegNet only) &  93.30 \% & 96.24 \% & 90.54 \% \\
3D (Depth only) &  88.42 \% & 86.83 \% & 90.07 \% \\
3D-2D (Ours) & 97.38 \% & 96.26 \% & 98.53 \% \\
 \bottomrule[1.5pt]
 \end{tabular}
\vspace{0.2cm}
  \caption{Quantitative Results on Camvid \cite{camvid} dataset} \label{tableCamvidtest}
\end{table}

Since our system requires a video and the KITTI UM/UMM datasets had unordered sets of labeled images which are not temporally connected, we could not use it. We took images at uniform intervals from 5 sequences($Seq02$, $Seq03$, $Seq05$, $Seq06$, $Seq08$) of the KITTI odometry dataset and labeled them into road and non-road using \cite{liblabel}. We labeled close to 150 images and evaluated the performance of our method using them against SegNet and depth based method as above. We will release the source code and annotations publicly when the manuscript is published.

On the Camvid dataset, our system outperforms both the compared methods in all three scores. While on KITTI odometry benchmark, we outperform on F val and recall but have a slightly lower precision than SegNet as shown in Table \ref{tableKITTItest}. This is because SegNet underestimates the road in KITTI dataset as it is not trained on it, which is also depicted by a very low recall. In comparison, our system has a much higher recall.

\begin{table}[htp]
\centering
\begin{tabular}{lccc}
\toprule[1.5pt]
\bf Method & \bf F val & \bf Precision & \bf Recall  \\ \hline
2D (SegNet only) &  85.07 \% & 90.49 \% & 80.25 \% \\
3D (Depth only) &  72.26 \% & 62.57 \% & 85.50 \% \\
3D-2D (Ours) & 89.22 \% & 87.10 \% & 91.45 \% \\
\bottomrule[1.5pt]
\end{tabular}
\vspace{0.2cm}
\caption{Quantitative Results on KITTI Odometry \cite{kitti} dataset} \label{tableKITTItest}
\end{table}

\subsection{Failure Cases}
\begin{figure}[!h]
\begin{center}
\includegraphics[width=\linewidth]{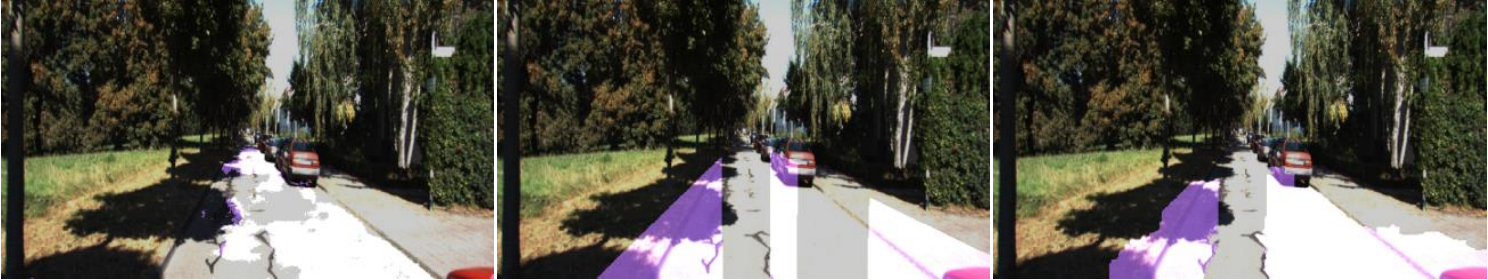}
\end{center}
\caption{Failure in free space detection due to inaccuracies in both 2D and 3D inputs. The first image shows segmentation in 2D using SegNet while the second image shows the projection of the inaccurate plane estimated by SLAM which leads to wrong free space estimation (third image). Please zoom in for better visualization.}
\label{fig:failure}
\end{figure}

Some of the possible failure cases for our method arise when both 2D segmentation and 3D information are incorrect. We analyse a failure example in Figure \ref{fig:failure}, where due to the inaccuracy of information from both 2D (first image) and 3D (second image) our method fails to correctly detect the free road space (third image).

\section{Conclusion}

In this paper, we have proposed a novel technique for free road space detection under varying illumination conditions using only a monocular camera. The proposed algorithm uses a joint 3D/2D based CRF formulation. We use SegNet for modelling road texture in 2D. We also use the sparse structure generated by SLAM to generate higher level priors which in-turn generates additional inferences from 3D about the road and obstacles. The 3D cues help in filling out gaps in the 2D road detection using texture and also corrects the inaccuracies occurring due to errors in segmentation. 2D priors, on the other hand, helps to correct areas where inaccuracies occur in estimated depth. Both of these cues complement each other to create a holistic model for free space detection on roads. In addition, we use color lines model for illumination invariance under different lighting conditions, which also helps as a smoothness parameter in case of unmarked roads. \\
{\bf Acknowledgement:} This work was supported by a research grant from Continental Automotive Components (India) Pvt. Ltd. Chetan Arora is supported by Visvesaraya Young Faculty Fellowship and Infosys Center for AI.

{\small
\bibliographystyle{ieee}
\bibliography{egbib}
}

\end{document}